\documentclass[conference]{IEEEtran}
\IEEEoverridecommandlockouts
\usepackage{cite}
\usepackage{amsmath,amssymb,amsfonts}
\usepackage{algorithmic}
\usepackage{graphicx}
\usepackage{textcomp}
\usepackage{xcolor}
\def\BibTeX{{\rm B\kern-.05em{\sc i\kern-.025em b}\kern-.08em
    T\kern-.1667em\lower.7ex\hbox{E}\kern-.125emX}}

\usepackage{graphicx}

\usepackage{tikz}
\usetikzlibrary{arrows.meta,positioning,fit,calc}
\usepackage{quantikz}
\usepackage{rotating} 
\usepackage{xcolor}
\definecolor{encblue}{RGB}{93, 157, 214}   
\definecolor{parorange}{RGB}{237, 156, 55} 
\definecolor{boxgreen}{RGB}{66, 148, 72}   
\definecolor{lightblue}{HTML}{E6E5FD}
\definecolor{lightgreen}{RGB}{196, 227, 190}

\usepackage{longtable}
\usepackage{pdflscape}
\usepackage{lineno}
\usepackage{booktabs}
\usepackage{tabularx}
\usepackage{placeins}
\usepackage[table]{xcolor}
\usepackage{comment}

\usepackage{hyperref}

\usepackage{url}
\usepackage{multirow}

\usepackage{amsmath,amssymb,bm}

\usepackage{graphicx}
\usepackage{tikz}
\usetikzlibrary{arrows.meta,positioning,fit,calc,decorations.pathreplacing}
\usepackage{quantikz}
\usepackage{xcolor}
\definecolor{encblue}{RGB}{93, 157, 214}
\definecolor{parorange}{RGB}{237, 156, 55}
\definecolor{boxgreen}{RGB}{66, 148, 72}
\usepackage{amsmath,amssymb,bm}

\usepackage[T1]{fontenc}
\usepackage{lmodern}
\usepackage{amsmath,amssymb}
\usepackage{tikz}
\usetikzlibrary{arrows.meta,calc,positioning,fit,backgrounds}

\usepackage{amsmath,amssymb}
\usepackage{tikz}
\usetikzlibrary{arrows.meta,calc,positioning,fit,backgrounds}

\usepackage{tikz}
\usetikzlibrary{arrows.meta,positioning,calc,shapes.geometric,decorations.pathmorphing,fit}

\begin{document}
\bstctlcite{IEEEexample:BSTcontrol}

\title{Hybrid Quantum Neural Network for Multivariate Clinical Time Series Forecasting}

\author{
\IEEEauthorblockN{
Irene Iele\IEEEauthorrefmark{1}, 
Floriano Caprio\IEEEauthorrefmark{1}, 
Paolo Soda\IEEEauthorrefmark{1}\IEEEauthorrefmark{2},
Matteo Tortora\IEEEauthorrefmark{3}
}

\IEEEauthorblockA{\IEEEauthorrefmark{1}
Unit of Artificial Intelligence and Computer Systems, Università Campus Bio-Medico di Roma, Italy}

\IEEEauthorblockA{\IEEEauthorrefmark{2}
Department of Diagnostics and Intervention, Radiation Physics,  
Biomedical Engineering, Umeå University, Sweden}

\IEEEauthorblockA{\IEEEauthorrefmark{3}
Department of Naval, Electrical, Electronics and Telecommunications Engineering,  
University of Genoa, Italy}

\thanks{Corresponding author: Matteo Tortora (matteo.tortora@unige.it)}
}

\maketitle

\begin{abstract}
Forecasting physiological signals can support proactive monitoring and timely clinical intervention by anticipating critical changes in patient status. 
In this work, we address multivariate multi-horizon forecasting of physiological time series by jointly predicting heart rate, oxygen saturation, pulse rate, and respiratory rate at forecasting horizons of 15, 30, and 60 seconds. We propose a hybrid quantum-classical architecture that integrates a Variational Quantum Circuit (VQC) within a recurrent neural backbone. 
A GRU encoder summarizes the historical observation window into a latent representation, which is then projected into quantum angles used to parameterize the VQC. 
The quantum layer acts as a learnable non-linear feature mixer, modeling cross-variable interactions before the final prediction stage. 
We evaluate the proposed approach on the BIDMC PPG and Respiration dataset under a Leave-One-Patient-Out protocol.
The results show competitive accuracy compared with classical and deep learning baselines, together with greater robustness to noise and missing inputs. 
These findings suggest that hybrid quantum layers can provide useful inductive biases for physiological time series forecasting in small-cohort clinical settings. 
The code is available at \url{https://github.com/arco-group/quantum-ml}.
\end{abstract}

\section{Introduction}
Continuous monitoring technologies generate dense physiological time series that can support proactive care~\cite{chromik2022alarmfatigue,coser2025deep}. 
Beyond retrospective risk assessment, forecasting vital sign trajectories can support early warning and clinical decision-making by anticipating clinically relevant changes before they become critical. 
The value of a forecast depends on its prediction horizon: short horizons can support rapid response and alarm management, whereas longer horizons can support earlier clinical planning.

In this work, we address multivariate multi-horizon forecasting of physiological time series, jointly predicting future values of heart rate (HR), oxygen saturation ($SpO_2$), pulse rate (Pulse), and respiratory rate (RR) at multiple forecasting horizons. 
More broadly, the joint modeling of multiple physiological streams can be viewed through a multimodal learning perspective, in which complementary information sources are integrated to capture dependencies that may be missed when each signal is modeled in isolation. Similar integration strategies have improved predictive performance across diverse domains by leveraging complementary information from multiple sources\cite{tortora2023radiopathomics,ayllon2025context,furia2023exploring,iele2026probabilistic,cordelli2024machine}.
In physiological monitoring, this perspective is particularly relevant because correlated vital signs provide distinct yet related views of patient status. 
Accordingly, multi-output forecasting is clinically meaningful for two reasons. 
First, multi-horizon prediction aligns with heterogeneous clinical decision time scales. 
Second, jointly modeling multiple signals can exploit physiological coupling and reduce inconsistencies that may arise when each variable is predicted independently~\cite{saleh2025streamhealth}.

Despite this motivation, reliable forecasting from clinical monitoring data remains challenging. 
Physiological time series are affected by sensor artifacts, missingness, and non-stationary dynamics induced by interventions and evolving patient conditions~\cite{furia2023exploring}. 
More broadly, health-related temporal processes often exhibit evolving dynamics over time, which complicates forecasting and motivates modeling approaches that can adapt to changing conditions~\cite{cordelli2020time}.
Generalization is further complicated by substantial inter-patient variability, particularly in small-cohort settings. 
In this regime, evaluation protocols must be designed carefully, since extracting multiple windows from the same patient can introduce subject leakage and yield overly optimistic estimates~\cite{joeres2025data}. 
For this reason, patient-independent evaluation is essential. 
Accordingly, we adopt a leave-one-patient-out (LOPO) protocol, which reflects a realistic deployment scenario in which forecasting models must generalize to unseen individuals.

In parallel, quantum machine learning (QML) has attracted growing interest in healthcare applications, although the empirical evidence remains preliminary. 
Recent reviews highlight recurring limitations, including weak benchmarking, limited evaluation under realistic operating conditions, and inconsistent superiority over classical approaches~\cite{gupta2025systematic}. 
This suggests a cautious perspective. 
Rather than assuming quantum advantage, QML components should be studied as structured inductive biases within hybrid pipelines and evaluated against strong classical baselines under realistic protocols.

Motivated by these considerations, we investigate a hybrid quantum-classical architecture for multivariate physiological forecasting. 
Our model integrates a Variational Quantum Circuit (VQC) into a compact recurrent backbone, using the quantum module as a learnable non-linear feature mixer for multivariate temporal representations. 
We evaluate this approach against strong classical baselines and deep learning models designed for short-horizon physiological forecasting.

Our main contributions are as follows:
\begin{itemize}
    \item We formulate physiological prediction as a multivariate multi-horizon forecasting task, jointly predicting HR, $SpO_2$, Pulse, and RR at horizons of \(h \in \{15,30,60\}\) seconds.
    \item We propose a hybrid quantum-classical forecasting model that incorporates a VQC layer as a learnable feature-mixing module for multivariate physiological time series.
    \item We provide a rigorous patient-independent evaluation under LOPO against classical and deep learning models, reporting RMSE and MAE across targets and horizons.
    \item We conduct ablation studies on the quantum module and key design choices to assess the contribution and limitations of the quantum component.

\end{itemize}

\begin{figure*}
    \centering
\includegraphics[width=0.75\linewidth]{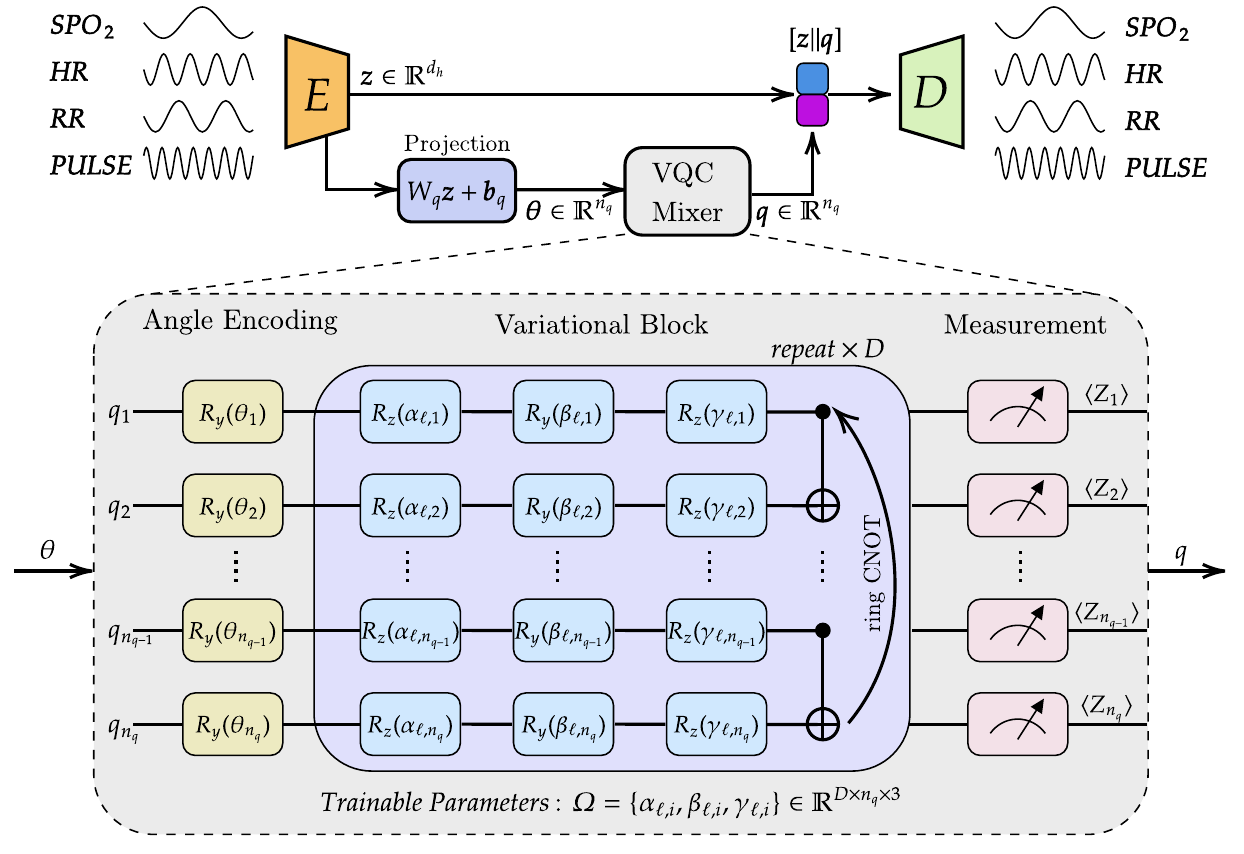}
\caption{Proposed hybrid GRU--VQC forecaster. The encoder \(E\) maps the input window to \(z\), projected to VQC angles \(\theta\). The VQC acts as a learnable non-linear feature mixer and outputs quantum features \(q\) via Pauli-\(Z\) measurements. The hybrid vector \([z\|q]\) is used to predict HR, \(SpO_2\), Pulse, and RR at \(h\in\{15,30,60\}\) s.}
    \label{fig:vqc_repeatD}
\end{figure*}

\section{Methods}
Our method is a hybrid quantum-classical forecaster composed of a GRU encoder~\cite{chung2014empirical} and a Variational Quantum Circuit (VQC) used as a learnable quantum feature mixer.  
Given a multivariate input window \(X \in \mathbb{R}^{L \times d_x}\), where \(L\) is the number of historical time steps and \(d_x\) is the number of input channels, the GRU encoder maps \(X\) to a latent representation \(z \in \mathbb{R}^{d_z}\). 
A learnable affine projection then maps \(z\) to the quantum angles:
\begin{equation}
\theta = W_q z + b_q,
\label{eq:to_qubits}
\end{equation}
where \(W_q \in \mathbb{R}^{n_q \times d_z}\) and \(b_q \in \mathbb{R}^{n_q}\) are learnable parameters. 

As shown in~\autoref{fig:vqc_repeatD}, the VQC first applies angle encoding by using \(R_y(\theta_j)\) on each qubit \(j=1,\dots,n_q\).  
The encoded state is then processed by a variational circuit of depth \(D\). 
At repetition \(\ell=1,\dots,D\), trainable single-qubit rotations are applied to each qubit \(j\):
\begin{equation}
U_{\mathrm{rot}}^{(\ell)}(\Omega_\ell)
= \bigotimes_{j=1}^{n_q}
\Big(R_z(\alpha_{\ell,j})\,R_y(\beta_{\ell,j})\,R_z(\gamma_{\ell,j})\Big),
\end{equation}
where \(\Omega_\ell=\{(\alpha_{\ell,j},\beta_{\ell,j},\gamma_{\ell,j})\}_{j=1}^{n_q}\) denotes the trainable rotation angles at repetition \(\ell\), and \(\bigotimes\) denotes the Kronecker product. 
Entanglement is introduced through an operator \(\mathcal{E}\), which reduces to the identity when entanglement is disabled. 
We adopt a ring topology:
\begin{equation}
\mathcal{E}_{\mathrm{ring}}
=\Big(\prod_{j=1}^{n_q-1}\mathrm{CNOT}(j,j{+}1)\Big)\mathrm{CNOT}(n_q,1),
\label{eq:ring_ent}
\end{equation}
and collect all trainable angles as \(\Omega=\{\Omega_\ell\}_{\ell=1}^D\). 
The overall circuit unitary is:
\begin{equation}
U_{\mathrm{VQC}}(\theta,\Omega)
=\left(\prod_{\ell=1}^{D} \mathcal{E}\, U_{\mathrm{rot}}^{(\ell)}(\Omega_\ell)\right)\,U_{\mathrm{emb}}(\theta).
\label{eq:Uvqc}
\end{equation}
The quantum circuit is initialized in the computational basis state \(\lvert 0\rangle^{\otimes n_q}\).
The quantum readout is defined through the measurement map \(\mathcal{M}_Z\), which returns the vector of Pauli-\(Z\) expectation values:
\begin{equation}
q = \mathcal{M}_Z\!\left(U_{\mathrm{VQC}}(\theta,\Omega)\lvert 0\rangle^{\otimes n_q}\right)
= \big[\langle Z_1\rangle,\ldots,\langle Z_{n_q}\rangle\big] \in \mathbb{R}^{n_q}.
\label{eq:q_readout}
\end{equation}

This construction can also be interpreted as a learnable quantum feature map. 
In quantum machine learning, parameterized circuits embed classical inputs into a quantum Hilbert space and induce nonlinear feature transformations. 
In our setting, the VQC receives the latent representation \(z\) produced by the GRU encoder and maps it, through angle encoding and variational processing, to the quantum feature vector \(q\). 
The circuit can therefore be viewed as a task-adaptive quantum feature map applied to the classical embedding, since its variational parameters \(\Omega\) are optimized jointly with the rest of the model during training.

The final hybrid representation is obtained by concatenating classical and quantum features, \(\tilde{z}=[z \,\|\, q]\). 
For each forecasting horizon \(h \in \{15,30,60\}\), predictions are produced through a linear head,
\begin{equation}
\hat{y}_h = W_h \tilde{z} + b_h,
\qquad h \in \{15,30,60\},
\end{equation}
where \(\hat{y}_h \in \mathbb{R}^{4}\) contains the predictions for HR, \(SpO_2\), Pulse, and RR at horizon \(h\).

\section{Experimental Configuration}
\subsection{Dataset}
We use the BIDMC PPG and Respiration Dataset~\cite{BIDMC}, which contains 53 ICU recordings of 8 minutes each.
For each subject, the dataset provides waveform signals sampled at 125~Hz and monitor-derived numerical parameters sampled at 1~Hz.
In this study, we use only the per-subject numerical streams HR, \(SpO_2\), Pulse, RR at 1~Hz.

\subsection{Implementation details}
The input window length is set to \(L = 240\) s, with signals sampled at 1 Hz, so that each model observes 4 minutes of history to predict future horizons \(h \in \{15, 30, 60\}\) s. 
All models are trained for 30 epochs with batch size 128 using the Adam optimizer with learning rate \(10^{-3}\). 
The proposed model combines a GRU backbone~\cite{chung2014empirical} with a VQC-based quantum mixer implemented in PennyLane~\cite{bergholm2018pennylane}. 
The circuit uses \(n_q=6\) qubits and depth \(D=3\), with entanglement enabled. 
Here, \(D\) denotes the number of repeated variational layers applied after angle encoding. 
The VQC is simulated in analytic mode, without finite-shot sampling. 
Model selection and evaluation follow a Leave-One-Patient-Out (LOPO) protocol over 53 folds, with one subject held out at each fold.

\subsection{Competitors}
We compare against forecasting baselines spanning multiple modeling paradigms, including tree-based ensembles, convolutional architectures, recurrent networks, and Transformer-based time series models. 
As a non-deep-learning baseline, we consider ExtraTrees (50 trees, maximum depth 2), which provides a strong non-linear reference in limited-data settings. 
Among convolutional approaches, we include a 1-D CNN (32 channels, kernel size 3, dropout 0.1), which captures local temporal patterns through stacked convolutions over the multivariate input sequence. 
We also evaluate a Temporal Convolutional Network (TCN)~\cite{bai2018empirical} (3 levels, 32 channels, kernel size 3, dropout 0.1), which uses dilated causal convolutions to model longer temporal dependencies while preserving causality. 
As a recurrent baseline, we adopt an LSTM~\cite{hochreiter1997long} (hidden size 56, 1 layer), a standard benchmark for multivariate sequence forecasting. 
To represent Transformer-based forecasting, we include PatchTST~\cite{nie2022time} (patch length 16, stride 8, embedding dimension 32, 4 heads, 1 layer, feedforward dimension 64), which models temporal interactions over patch embeddings using self-attention. 
All models are trained and evaluated under the same forecasting horizons and protocol for fair comparison.

\section{Results}

\begin{table*}[t!]
\caption{Per-task and per-horizon forecasting performance, reported as mean MAE and RMSE over 53 LOPO folds.  \colorbox{lightblue}{Blue} cells indicate the best-performing model within each physiological task (averaged across horizons), \colorbox{lightgreen}{green} cells indicate the best-performing model within each forecasting horizon (averaged across tasks), and \textbf{BOLD} values denote the best result for each task--horizon pair. The AvgWins row reports, for each method, the percentage of task--horizon pairs in which it achieves the best performance.}
\label{tab:results_task_horizon_models}
\centering
\resizebox{\textwidth}{!}{
\begin{tabular}{ll
cc cc cc cc cc cc cc}
\toprule

\multirow{2}{*}{\textbf{Task}} & \multirow{2}{*}{\textbf{H}}
& \multicolumn{2}{c}{\textbf{ExtraTrees}}
& \multicolumn{2}{c}{\textbf{CNN 1D}} 
& \multicolumn{2}{c}{\textbf{TCN}} 
& \multicolumn{2}{c}{\textbf{LSTM}} 
& \multicolumn{2}{c}{\textbf{PatchTST}} 
& \multicolumn{2}{c}{\textbf{GRU + VQC (Ours)}} \\
\cmidrule(lr){3-4}
\cmidrule(lr){5-6}
\cmidrule(lr){7-8}
\cmidrule(lr){9-10}
\cmidrule(lr){11-12}
\cmidrule(lr){13-14}
& & \textbf{MAE$\downarrow$}
  & \textbf{RMSE$\downarrow$} & \textbf{MAE$\downarrow$}
  & \textbf{RMSE$\downarrow$} & \textbf{MAE$\downarrow$}
  & \textbf{RMSE$\downarrow$} & \textbf{MAE$\downarrow$} 
  & \textbf{RMSE$\downarrow$} & \textbf{MAE$\downarrow$}
    & \textbf{RMSE$\downarrow$} & \textbf{MAE$\downarrow$}
        & \textbf{RMSE$\downarrow$} \\
\midrule

\multirow{4}{*}{HR}
& 15 & $.367_{\pm .291}$ & $.394_{\pm .298}$ & $.185_{\pm .095}$ & $.234_{\pm .123}$ & $.138_{\pm .099}$ & $.180_{\pm .132}$ & $\mathbf{.135}_{\pm \mathbf{.102}}$ & $\mathbf{.178}_{\pm \mathbf{.136}}$& $.166_{\pm .151}$ & $.202_{\pm .158}$ & $.140_{\pm .097}$ & $.179_{\pm .125}$\\

& 30 & $.363_{\pm .303}$ & $.390_{\pm .302}$ & $.221_{\pm .116}$ & $.275_{\pm .143}$ & $.154_{\pm .104}$ & $.198_{\pm .133}$ & $.154_{\pm .111}$ & $.198_{\pm .139}$ & $.175_{\pm .162}$ & $.212_{\pm .170}$ & $\mathbf{.148}_{\pm \mathbf{.103}}$ & $\mathbf{.190}_{\pm \mathbf{.129}}$\\

& 60 & $.374_{\pm .307}$ & $.403_{\pm .305}$ 
& $.231_{\pm .111}$ & $.286_{\pm .137}$ & $.164_{\pm .105}$ & $.208_{\pm .142}$ & $.155_{\pm .105}$ & $.200_{\pm .140}$ & $.178_{\pm .151}$ & $.219_{\pm .167}$ & $\mathbf{.161}_{\pm \mathbf{.110}}$ & $\mathbf{.205}_{\pm \mathbf{.140}}$\\

& \textbf{Avg} & $.368_{\pm .299}$ & $.396_{\pm .299}$ & $.212_{\pm .100}$ & $.265_{\pm .127}$ & $.152_{\pm .098}$ & $.195_{\pm .131}$ &\cellcolor{lightblue} $.148_{\pm .101}$ & $.192_{\pm .134}$ & $.173_{\pm .152}$ & $.211_{\pm .161}$ & $.150_{\pm.100}$ 
&\cellcolor{lightblue} 
$.191_{\pm.127}$ \\
\midrule

\multirow{4}{*}{SpO$_2$}
& 15 & $.247_{\pm .124}$ & $.274_{\pm .134}$ & $.197_{\pm .106}$ & $.245_{\pm .128}$ & $.136_{\pm .100}$ & $.182_{\pm .125}$ & $.129_{\pm .109}$ & $.177_{\pm .135}$ & $.170_{\pm .163}$ & $.205_{\pm .168}$ & $\mathbf{.127}_{\pm \mathbf{.099}}$ & $\mathbf{.171}_{\pm \mathbf{.126}}$\\

& 30 & $.245_{\pm .131}$ & $.271_{\pm .142}$ & $.213_{\pm .096}$ & $.262_{\pm .115}$ & $.143_{\pm .092}$ & $.188_{\pm .111}$ & $.128_{\pm .090}$ & $.176_{\pm .114}$ & $.172_{\pm .159}$ & $.203_{\pm .164}$ & $\mathbf{.125}_{\pm \mathbf{.088}}$ & $\mathbf{.173}_{\pm \mathbf{.114}}$ \\

& 60 & $.249_{\pm .139}$ & $.276_{\pm .153}$
& $.236_{\pm .106}$ & $.289_{\pm .131}$ & $.152_{\pm .098}$ & $.199_{\pm .117}$ & $.144_{\pm .104}$ & $.193_{\pm .128}$ & $.181_{\pm .167}$ & $.211_{\pm .170}$ & $\mathbf{.146}_{\pm \mathbf{.093}}$ & $\mathbf{.191}_{\pm \mathbf{.118}}$\\

& \textbf{Avg} & $.247_{\pm .127}$ & $.274_{\pm .138}$ & $.215_{\pm .098}$ & $.265_{\pm .120}$ & $.144_{\pm .092}$ & $.190_{\pm .113}$ & $.134_{\pm .098}$ & $.182_{\pm .123}$ & $.174_{\pm .161}$ & $.206_{\pm .166}$ &\cellcolor{lightblue} 
$.133{\pm .090}$ 
&\cellcolor{lightblue} 
$.178{\pm .116}$ \\
\midrule

\multirow{4}{*}{Pulse}
& 15 & $.331_{\pm .291}$ & $.282_{\pm .353}$ & $.177_{\pm .108}$ & $.222_{\pm .130}$ & $.123_{\pm .095}$ & $.153_{\pm .116}$ & $.125_{\pm .096}$ & $.157_{\pm .119}$ & $.160_{\pm .134}$ & $.185_{\pm .139}$ & $\mathbf{.116}_{\pm \mathbf{.091}}$ & $\mathbf{.146}_{\pm \mathbf{.112}}$\\

& 30 & $.287_{\pm .361}$ & $.286_{\pm .561}$ & $.201_{\pm .111}$ & $.248_{\pm .136}$ & $.141_{\pm .109}$ & $.175_{\pm .131}$ & $.135_{\pm .110}$ & $.169_{\pm .134}$ & $.168_{\pm .136}$ & $.193_{\pm .142}$ & $\mathbf{.137}_{\pm \mathbf{.116}}$ & $\mathbf{.169}_{\pm \mathbf{.136}}$ \\

& 60 & $.342_{\pm .287}$ & $.367_{\pm .287}$ & $.222_{\pm .120}$ & $.271_{\pm .147}$ & $.147_{\pm .106}$ & $.182_{\pm .132}$ & $.136_{\pm .103}$ & $.172_{\pm .130}$ & $.177_{\pm .153}$ & $.204_{\pm .162}$ & $\mathbf{.140}_{\pm \mathbf{.106}}$ & $\mathbf{.173}_{\pm \mathbf{.129}}$ \\

& \textbf{Avg} & $.337_{\pm .283}$ & $.361_{\pm .283}$ & $.200_{\pm .106}$ & $.247_{\pm .131}$ & $.137_{\pm .099}$ & $.170_{\pm .122}$ & $.132_{\pm .101}$ & $.166_{\pm .125}$ & $.168_{\pm .139}$ & $.194_{\pm .145}$ &\cellcolor{lightblue} 
$.131{\pm .102}$ 
&\cellcolor{lightblue} 
$.162{\pm .123}$ \\
\midrule

\multirow{4}{*}{RR}
& 15 & $.455_{\pm .513}$ & $.529_{\pm .532}$ & $.416_{\pm .244}$ & $.519_{\pm .309}$ & $.339_{\pm .223}$ & $.421_{\pm .289}$ & $.320_{\pm .254}$ & $.412_{\pm .324}$ & $.403_{\pm .464}$ & $.491_{\pm .497}$ & $\mathbf{.311}_{\pm \mathbf{.225}}$ & $\mathbf{.389}_{\pm \mathbf{.297}}$\\

& 30 & $.561_{\pm .513}$ & $.644_{\pm .542}$ & $.540_{\pm .361}$ & $.648_{\pm .407}$ & $.476_{\pm .332}$ & $.565_{\pm .387}$ & $.458_{\pm .392}$ & $.556_{\pm .452}$ & $.432_{\pm .504}$ & $.516_{\pm .537}$ & $\mathbf{.421}_{\pm \mathbf{.355}}$ & $\mathbf{.505}_{\pm \mathbf{.410}}$\\

& 60 & $.582_{\pm .559}$ & $.667_{\pm .583}$ 
& $.521_{\pm .319}$ & $.645_{\pm .395}$ & $.540_{\pm .401}$ & $.632_{\pm .448}$ & $.494_{\pm .433}$ & $.596_{\pm .487}$ & $.456_{\pm .529}$ & $.546_{\pm .564}$ & $\mathbf{.466}_{\pm \mathbf{.466}}$ & $\mathbf{.560}_{\pm \mathbf{.515}}$\\

& \textbf{Avg} & $.533_{\pm .526}$ & $.614_{\pm .549}$ & $.492_{\pm .297}$ & $.604_{\pm .361}$ & $.452_{\pm .314}$ & $.539_{\pm .370}$ & $.424_{\pm .355}$ & $.522_{\pm .416}$ & $.430_{\pm .497}$ & $.517_{\pm .530}$ &\cellcolor{lightblue} 
$.399{\pm.345}$
&\cellcolor{lightblue} $.485_{\pm.405}$\\
\midrule

\multicolumn{2}{c}{{\textbf{AvgWins}}} & \multicolumn{2}{c}{{\textbf{0\%}}} & \multicolumn{2}{c}{{\textbf{0\%}}}  & \multicolumn{2}{c}{{\textbf{0\%}}}  & \multicolumn{2}{c}{{\textbf{8.4\%}}} & \multicolumn{2}{c}{{\textbf{0}}}  & \multicolumn{2}{c}{{\textbf{91.6 \%}}}  \\
\hline
\multirow{4}{*}{{Avg}}
& 15 & $.349_{\pm .199}$ & $.389_{\pm .199}$
& $.244_{\pm .108}$ & $.305_{\pm .134}$ & $.184_{\pm .084}$ & $.234_{\pm .104}$ & $.177_{\pm .092}$ & $.231_{\pm .111}$ & $.225_{\pm .166}$ & $.270_{\pm .172}$ &\cellcolor{lightgreen}
$.173_{\pm .081}$ 
&\cellcolor{lightgreen}$.221_{\pm .102}$\\

& 30 & $.377_{\pm .198}$ & $.417_{\pm .200}$ 
& $.294_{\pm .140}$ & $.358_{\pm .161}$ & $.229_{\pm .114}$ & $.281_{\pm .131}$ & $.219_{\pm .124}$ & $.275_{\pm .141}$ & $.237_{\pm .182}$ & $.281_{\pm .188}$ &\cellcolor{lightgreen} $.208_{\pm .115}$ &\cellcolor{lightgreen}$.259_{\pm .132}$\\

& 60 & $.387_{\pm .208}$ & $.428_{\pm .211}$
& $.302_{\pm .124}$ & $.373_{\pm .151}$ & $.251_{\pm .119}$ & $.305_{\pm .137}$ & $.232_{\pm .126}$ & $.290_{\pm .144}$ & $.248_{\pm .188}$ & $.295_{\pm .195}$ &\cellcolor{lightgreen}$.228_{\pm .142}$ &\cellcolor{lightgreen} $.282_{\pm .158}$\\

&\textbf{Avg} & $.371_{\pm .201}$ & $.411_{\pm .203}$ & $.280_{\pm .121}$ & $.345_{\pm .146}$ & $.221_{\pm .103}$ & $.274_{\pm .121}$ & $.209_{\pm .112}$ & $.265_{\pm .129}$ & $.237_{\pm .178}$ & $.282_{\pm .184}$ &
$.203_{\pm .110}$ & $.254_{\pm .128}$\\
\bottomrule
\end{tabular}
}
\end{table*}

All experiments are conducted under the Leave-One-Patient-Out (LOPO) protocol with $N=53$ test subjects.
We report standard point-forecast errors, MAE and RMSE~\cite{hyndman2006another,tortora2023matnet}, for each target and forecast horizon.
To summarize performance, we compute: (i) per-target averages across horizons, (ii) per-horizon averages across targets, and (iii) a macro-average across all target--horizon pairs.

\subsection{Forecasting task}
\autoref{tab:results_task_horizon_models} summarizes the forecasting performance of all methods across physiological targets and prediction horizons. 
Blue cells denote the best-performing model for each task after averaging across horizons, whereas green cells denote the best-performing model for each horizon after averaging across tasks. 
Bold values indicate the best result for each task-horizon pair. 
Each block of rows corresponds to one physiological signal (HR, \(SpO_2\), Pulse, RR), with results reported for \(h \in \{15,30,60\}\) s together with the average across horizons; the final block reports averages aggregated across all tasks.

The proposed GRU+VQC model achieves the strongest overall performance. 
It attains the highest AvgWins score at 91.6\%, where AvgWins denotes the percentage of task-horizon pairs in which a method achieves the lowest error among the compared models. 
This advantage is also reflected in the aggregated results, where GRU+VQC achieves the lowest MAE and RMSE at all forecasting horizons and in the overall average across tasks and horizons. 
At the task level, the proposed model achieves the best average performance for \(SpO_2\), Pulse, and RR,  while remaining very close to the best competitor on HR.

Among the baselines, ExtraTrees is clearly the weakest approach, whereas CNN 1D and TCN provide intermediate performance. 
LSTM and PatchTST emerge as the strongest classical competitors, with LSTM being especially competitive on HR and PatchTST showing solid overall behavior; however, both are generally surpassed by the proposed hybrid model.

As expected, forecasting errors increase with longer horizons for all methods. 
Importantly, the relative advantage of GRU+VQC is preserved from 15\,s to 60\,s, indicating that the proposed quantum-enhanced mixer provides useful cross-variable interactions for both short-term and longer-horizon forecasting

\subsection{Ablation study}
\begin{figure}
    \centering
\includegraphics[width=\columnwidth]{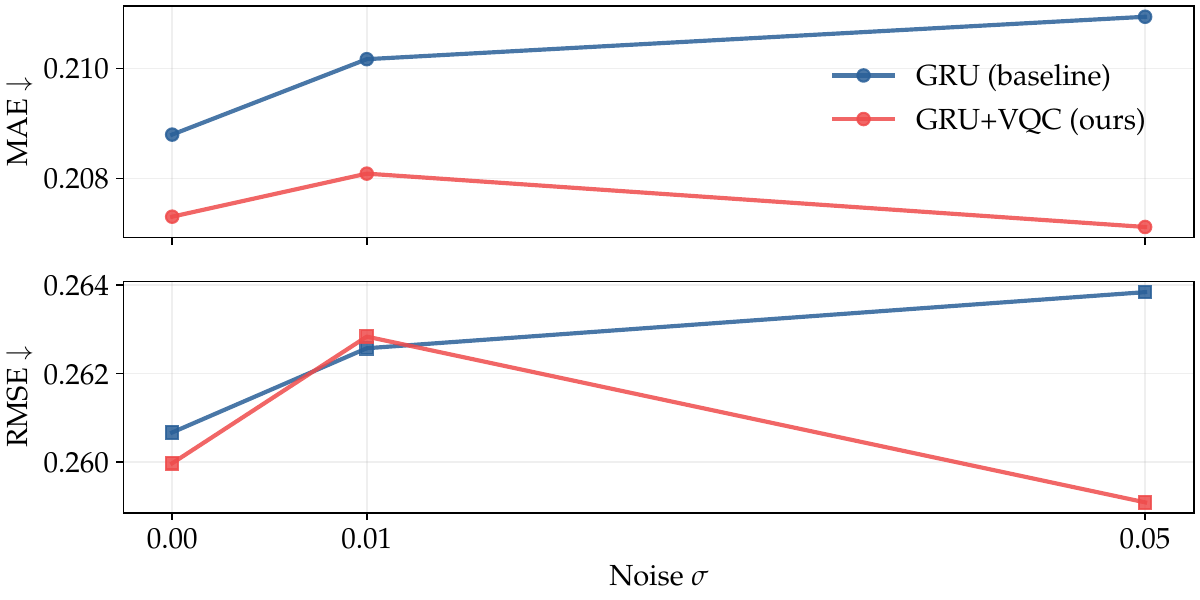}
\caption{Noise sensitivity analysis under increasing Gaussian perturbations applied to the test inputs. Performance is reported in terms of macro-MAE and macro-RMSE under the LOPO protocol ($N=53$).}
    \label{fig:robustness_noise}
\end{figure}

\paragraph{Robustness to input noise}
We evaluate the robustness of the proposed model under additive Gaussian perturbations applied to the inputs at test time. 
The comparison involves the GRU baseline and the proposed GRU+VQC variant. 
Noise is injected only at test time, while training and validation data remain unchanged. 
Given an input window \(x\), after fold-specific standardization, the perturbed input is defined as
\[
x' = x + \epsilon, \qquad \epsilon \sim \mathcal{N}(0,\sigma^2),
\]
where \(\sigma\) controls the perturbation level. 
Noise is sampled independently for each time step and feature. 
We consider three perturbation levels, \(\sigma \in \{0.00, 0.01, 0.05\}\). 
For each noise level and LOPO fold, we compute macro-averaged MAE and RMSE across the four physiological variables (HR, Pulse, RR, \(SpO_2\)). 
Reported results are averaged over the 53 test subjects. 
\autoref{fig:robustness_noise} shows the resulting trends, with MAE in the upper panel and RMSE in the lower panel.

GRU+VQC shows a consistent robustness advantage across all perturbation levels. 
It achieves lower MAE and RMSE than the GRU baseline for all tested values of \(\sigma\). 
Whereas the baseline degrades monotonically as the perturbation level increases, GRU+VQC remains comparatively stable, with only limited variation across noise levels. 
These results suggest that the quantum mixer provides a regularizing effect under noisy inputs.

\begin{figure}
    \centering
\includegraphics[width=\columnwidth]{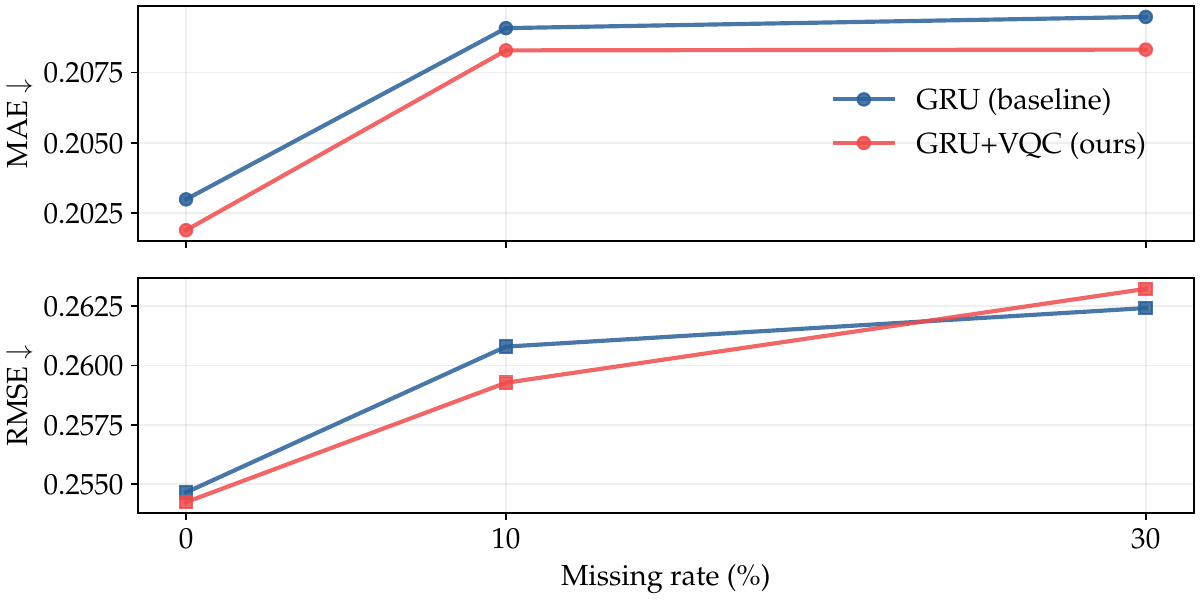}
\caption{Missing-data sensitivity analysis under increasing missing rates applied to the test inputs. 
Performance is reported in terms of macro-MAE and macro-RMSE under the LOPO protocol ($N=53$).}
    \label{fig:robustness_miss}
\end{figure}

\paragraph{Robustness to missing values}
We further assess robustness when test inputs contain missing values. 
Missingness is simulated by randomly masking input elements with probability \(p\), followed by forward-backward imputation, while keeping targets unchanged. 
We consider three missing rates, \(p \in \{0.0, 0.1, 0.3\}\). 
\autoref{fig:robustness_miss} reports the resulting trends. 
Both models degrade as the missing rate increases. 
GRU+VQC consistently achieves lower MAE than the GRU baseline across all missing rates. 
For RMSE, GRU+VQC performs slightly better at low and moderate missingness, while the two models become very close at the highest missing rate, where the GRU baseline is marginally better. 
These results indicate that the proposed hybrid model remains competitive under missing inputs, achieving consistently lower MAE while maintaining comparable RMSE.

\begin{figure*}[t]
    \centering
\includegraphics[width=0.85\textwidth]{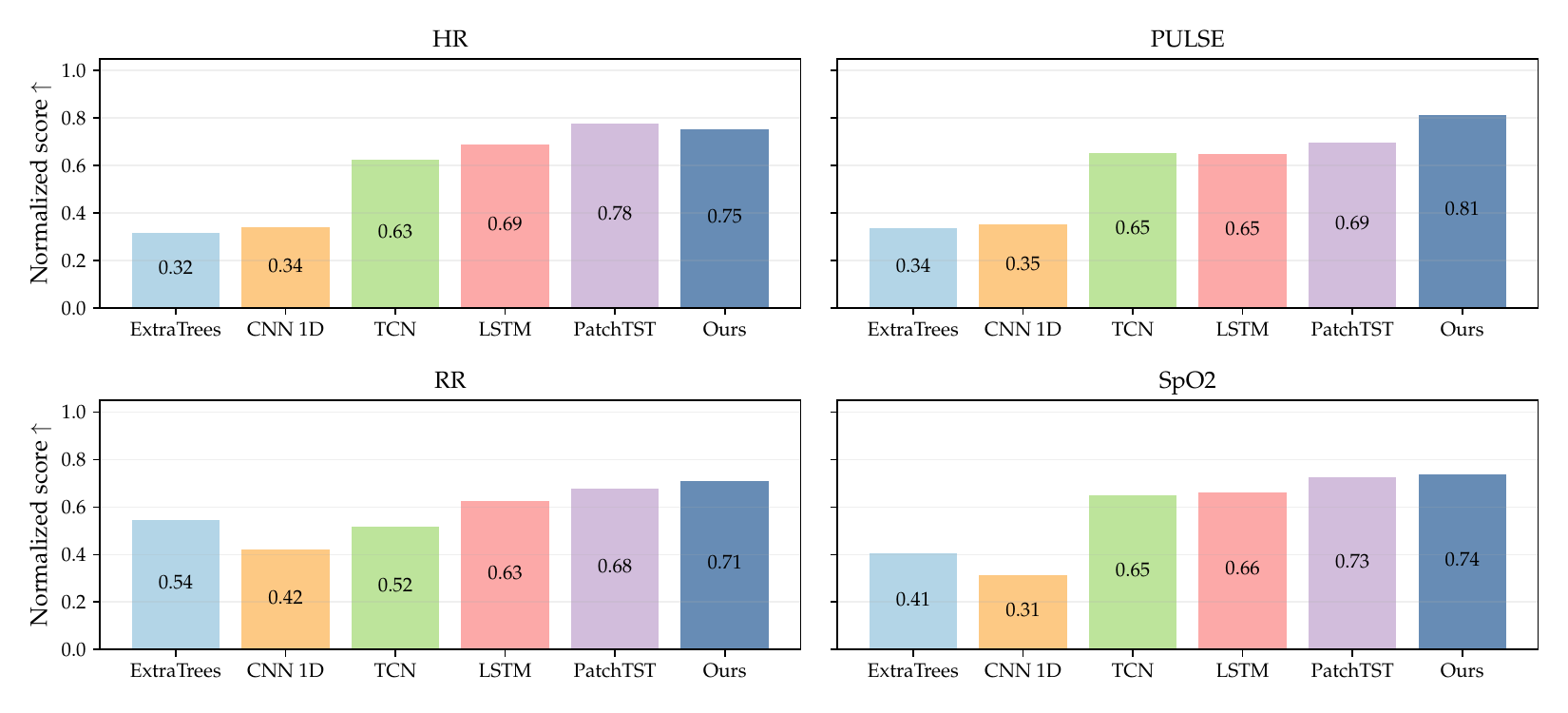}
\caption{Patient-wise ranking comparison across physiological tasks under the LOPO protocol ($N=53$). 
For each subject, RMSE is first averaged across prediction horizons ($h15$, $h30$, $h60$), models are ranked by increasing error, and scores are normalized to $[0,1]$.}
\label{fig:ours_competitor_by_task}
\end{figure*}

\begin{figure*}[t]
    \centering
\includegraphics[width=0.75\textwidth]{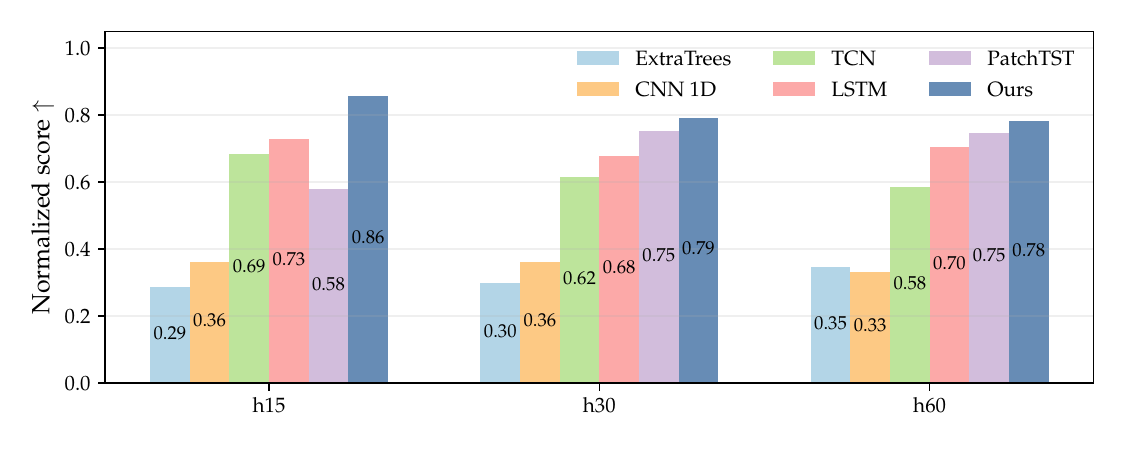}
\caption{Patient-wise ranking comparison across prediction horizons ($h15$, $h30$, $h60$). 
For each subject, RMSE is averaged across the four physiological tasks (HR, Pulse, RR, SpO$_2$), models are ranked by increasing error, and scores are normalized to $[0,1]$.}
\label{fig:ours_competitor_by_hor}
\end{figure*}

\subsection{Patient-wise ranking evaluation}
To complement the global error metrics, we evaluate the models using a patient-wise ranking procedure.
This analysis emphasizes the consistency of model performance across individual subjects rather than relying solely on the average error over the entire dataset.

For each subject \(s\), model \(m\), and physiological task \(t\), the prediction error is first aggregated across the three forecasting horizons \(h \in \{15,30,60\}\) using RMSE:
\begin{equation}
\mathrm{RMSE}_{t}^{(s,m)} =
\frac{1}{3}
\sum_{h \in \{\mathrm{15},\mathrm{30},\mathrm{60}\}}
\mathrm{RMSE}_{h,t}^{(s,m)} .
\end{equation}
For each subject and task, the six models are then ranked by increasing \(\mathrm{RMSE}_{t}^{(s,m)}\), and a score \(r \in \{6,5,4,3,2,1\}\) is assigned, with higher values indicating better ranks. 
Let \(score_{s,m}\) denote the score assigned to model \(m\) for subject \(s\). 
The final normalized ranking score [0,1] is computed as
\begin{equation}
\mathrm{Score}_{\mathrm{norm}}^{(m)} =
\frac{1}{6N}\sum_{s=1}^{N} score_{s,t,m} 
\end{equation}
where \(N=53\) is the number of test subjects and \(6\) is the maximum achievable rank. 
This score summarizes the average relative position of each model across patients and is less sensitive to subject-specific outliers than global mean error metrics.

We derive two complementary views from this procedure. 
First, task-wise performance is obtained by aggregating errors across horizons, yielding one ranking per physiological variable. 
Second, horizon-wise performance is obtained by aggregating errors across tasks. 
In the latter case, for each subject \(s\), model \(m\), and horizon \(h\), the prediction error is defined as
\begin{equation}
\mathrm{RMSE}_{h}^{(s,m)} =
\frac{1}{4}
\sum_{t \in \{\mathrm{HR},\mathrm{Pulse},\mathrm{RR},\mathrm{SpO_2}\}}
\mathrm{RMSE}_{h,t}^{(s,m)} .
\end{equation}
The same subject-wise ranking and normalization procedure is then applied.

Accordingly, \autoref{fig:ours_competitor_by_task} reports the task-wise comparison obtained by averaging across horizons, whereas \autoref{fig:ours_competitor_by_hor} reports the complementary horizon-wise comparison obtained by averaging across tasks.

\section{Conclusion}
We investigated a hybrid quantum-classical architecture for multivariate multi-horizon forecasting of physiological signals. 
The proposed model integrates a VQC as a learnable non-linear feature-mixing module within a GRU backbone. 
Results show competitive performance against classical baselines and deep learning models, together with improved robustness to noise and missing inputs. 
At the same time, the present study should be regarded as an exploratory analysis of quantum feature mixing, as it is conducted on a relatively small cohort and evaluated on a single dataset. 
Moreover, the proposed hybrid architecture is assessed using noiseless quantum circuit simulations. 
In this setting, the VQC is primarily used as a non-linear feature transformation within a hybrid learning pipeline, allowing its representational contribution to be studied independently of hardware limitations. 
Future work will extend the evaluation to larger and more diverse physiological datasets, and will investigate robustness under realistic noisy intermediate-scale quantum conditions, including gate errors, decoherence, and measurement noise. 
This will help clarify the practical viability of hybrid quantum-classical forecasting models on emerging quantum hardware.

\section*{Acknowledgment}
Irene Iele is a Ph.D. student enrolled in the National Ph.D. in Artificial Intelligence, course on Health and Life Sciences, organized by Università Campus Bio-Medico di Roma. 

\bibliographystyle{IEEEtran}
\bibliography{IEEEabrv,references}

\end{document}